\documentclass[runningheads]{llncs}
\usepackage{graphicx}

\usepackage{tikz}
\usepackage{comment}
\usepackage{amsmath,amssymb} 
\usepackage{color}
\usepackage{orcidlink}

\usepackage[accsupp]{axessibility}  

\usepackage{microtype}
\usepackage{color}
\usepackage{colortbl}
\usepackage{multirow}
\usepackage{makecell}
\usepackage{hhline}
\usepackage{cite}
\usepackage{caption}
\usepackage[american]{babel}
\usepackage{bbding}
\usepackage{cuted}
\definecolor{lightgray}{gray}{.92}
\definecolor{tinygray}{gray}{.96}

\newcommand{\etal}{\textit{et al}.}
\newcommand{\ie}{\textit{i}.\textit{e}.}
\newcommand{\eg}{\textit{e}.\textit{g}.}
\newcommand{\etc}{\textit{etc}}

\begin{document}
\pagestyle{headings}
\mainmatter
\def\ECCVSubNumber{2925}  

\title{From Face to Natural Image: Learning Real Degradation for Blind Image Super-Resolution} 

\titlerunning{Learning Real Degradation from Face to Natural Images}
\author{Xiaoming Li\inst{1,5}\orcidlink{0000-0003-3844-9308}\index{Xiaoming Li}
	\and
	Chaofeng Chen\inst{2}\orcidlink{0000-0001-6137-5162} 
	\and
	Xianhui Lin\inst{3} \and \\ 
	Wangmeng Zuo\inst{1,4(}\Envelope$^{)}$\orcidlink{0000-0002-3330-783X} 
	\and
	Lei~Zhang\inst{5}\orcidlink{0000-0002-2078-4215}}
\authorrunning{Xiaoming Li, et al.}
\institute{Faculty of Computing, Harbin Institute of Technology, China \and
	S-Lab, Nanyang Technological University, Singapore \and
	DAMO Academy, Alibaba Group, Shenzhen, China \and
	Peng Cheng Lab, Shenzhen, China \and
	Department of Computing, The Hong Kong Polytechnic University
	\\
	\email{\{csxmli, chaofenghust, xhlin129\}@gmail.com, wmzuo@hit.edu.cn, cslzhang@comp.polyu.edu.hk}
}

\maketitle

\begin{abstract}
How to design proper training pairs is critical for super-resolving real-world low-quality (LQ) images, which suffers from the difficulties in either acquiring paired ground-truth high-quality (HQ) images or synthesizing photo-realistic degraded LQ observations. 
Recent works mainly focus on modeling the degradation with handcrafted or estimated degradation parameters, which are however incapable to model complicated real-world degradation types, resulting in limited quality improvement. Notably, LQ face images, which may have the same degradation process as natural images, can be robustly restored with photo-realistic textures by exploiting their strong structural priors.
This motivates us to use the real-world LQ face images and their restored HQ counterparts to model the complex real-world degradation (namely ReDegNet), and then transfer it to HQ natural images to synthesize their realistic LQ counterparts. 
By taking these paired HQ-LQ face images as inputs to explicitly predict the degradation-aware and content-independent representations, we could control the degraded image generation, and subsequently transfer these degradation representations from face to natural images to synthesize the degraded LQ natural images.  
Experiments show that our ReDegNet can well learn the real degradation process from face images. The restoration network trained with our synthetic pairs performs favorably against SOTAs. 
More importantly, our method provides a new way to handle the real-world complex scenarios by learning their degradation representations from the facial portions, which can be used to significantly improve the quality of non-facial areas. The source code is available at \url{https://github.com/csxmli2016/ReDegNet}.
\keywords{real world degradation, blind image super-resolution}
\end{abstract}

\section{Introduction}
\label{sec:intro}
\begin{figure*}[!t]
	\centering
	\includegraphics[width=.92\textwidth]{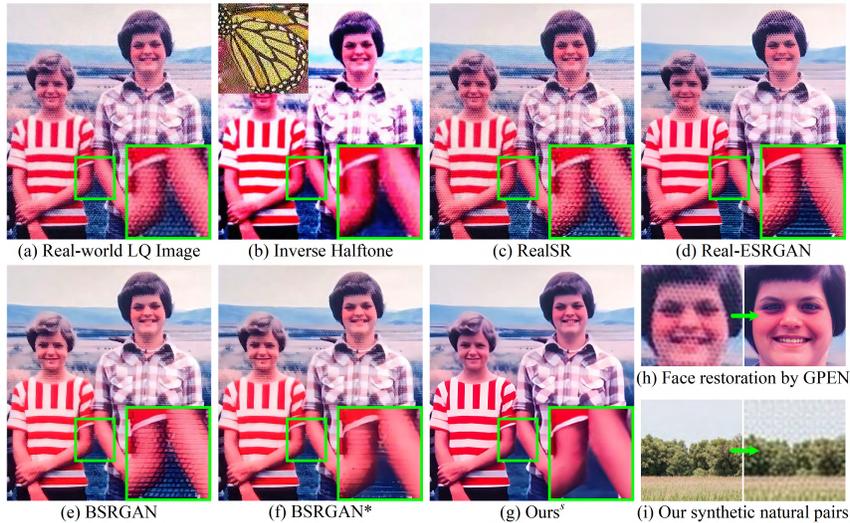} 
	\caption{(a): A real-world LQ image. (b)$\sim$(g): Restoration comparisons with inverse halftone method~\cite{son2020inverse}, RealSR~\cite{ji2020real}, Real-ESRGAN~\cite{wang2021realesrgan}, BSRGAN~\cite{zhang2021designing}, BSRGAN* fine-tuned with halftone degradation~\cite{gao2019deep}, and Ours$^s$ that is specifically trained with the synthetic pairs in (i). (h): Face restoration result by GPEN~\cite{Yang2021GPEN}. 
	(i): Our synthetic LQ sample with the degradation representation from (h).  
	}
	\label{fig:fig1}
\end{figure*}

It is widely known that Convolutional Neural Networks (CNNs) are proficient in handling the data they have seen, but perform inferior on these deviating from the training sets. 
This property makes the blind image super-resolution  networks difficult to handle the real-world LQ images which are usually corrupted with complex and unsynthesizable degradation.
However, building these pairs of real-world LQ and HQ datasets is neither feasible nor practical, because the real-world degradation types are too diverse and some of them are not brought by the imaging system.
Figure~\ref{fig:fig1} (a) shows a real-world LQ image that is degraded with halftone related artifacts. One can see that the synthetic LQ image  (on the top-left of (b)) by the inverse halftoning method~\cite{gao2019deep}  is hardly consistent with the complex real-world degradation, which makes these types of restoration methods (\eg, \cite{son2020inverse}) fail to generate photo-realistic result (see (b)). 

To alleviate the difficulties in restoring the real-world LQ images, some works attempt to predict the degradation parameters~\cite{guo2019toward,ji2020real,gu2019blind,luo2020unfolding,jiang2021towards} and then handle the LQ input with the non-blind restoration works. However, the real degradation usually combines with various corruption types, each of which has lost its intrinsic characteristics. This inevitably makes these methods sensitive to the prediction errors of the degradation parameters, and consequently makes them fail to handle the real-world LQ image (see (c) in Figure~\ref{fig:fig1}).

Recently, data-driven methods are suggested to design a practical degradation model by handcrafting the complex combinations of blur, downsampling, noise and JPEG compression with random~\cite{zhang2021designing} or high orders~\cite{wang2021realesrgan}. 
Albeit these methods have more diverse degradation types~\cite{elad1997restoration,liu2013bayesian,Zhang_2019_CVPR} and show great generalization in handling the real-world LQ images in most cases, 
they still fail to cover some complex real degradation which cannot be well synthesized 
(see (d) and (e) in Figure~\ref{fig:fig1}). 
By incorporating the synthetic halftone degradation~\cite{gao2019deep}, BSRGAN* has slight improvement (see (f)), but 
still contains obvious linearity artifacts.

In contrast, face image has specific and strong structure prior, and can be better restored while exhibiting great generalization ability on real-world LQ images in most cases~\cite{Li_2020_ECCV,wang2021towards,Yang2021GPEN}. 
Although the image is corrupted by intractable degradation, the face restoration result is very plausible and photo-realistic (see Figure~\ref{fig:fig1} (h)). Since the face and non-face (natural) regions in an image share the same degradation, once we have known the degradation process on face regions, transferring it to natural HQ images would bring considerable benefits, \eg, we can apply this degradation process on the HQ natural image to synthesize these types of natural image pairs (see (i)) for training restoration network (see (g)).

In this paper, we make the first attempt to explore the \textbf{re}al \textbf{deg}radation with ReDegNet, which contains
(i) learning the real degradation from the pairs of real-world LQ and pseudo HQ face images with DegNet, and (ii) transferring it to HQ natural images to synthesizing their realistic LQ ones with SynNet. 
As for (i), instead of taking a single LQ image to predict its degradation parameters~\cite{ji2020real}, our DegNet takes the real-world LQ and its pseudo HQ face images as input to generate the degradation representation, which models the degradation process of how the HQ image is degraded to the LQ one. 
To disentangle the image content and degradation type, we adopt two manners, \ie, 
a) carefully designed framework by predicting the degradation representation through several fully connected layers  to generate the convolution weights which can be regarded as the styles in StyleGANs~\cite{karras2019style,karras2020analyzing}, and b) contrastive loss~\cite{wang2021unsupervised} by minimizing the representation distance between the pairs with different content but degraded with the same degradation parameters, and meanwhile maximizing these with the same content but different degradation. 
This process is fully supervised by the paired LQ/HQ face images. 
As for (ii), our SynNet synthesizes the realistic LQ natural images with these degradation representations extracted from face images, which can help us to  learn the real-world restoration mapping. 
Note that our method may perform limited on scenarios without faces. By extending the degradation space with face images share the similar degradation, our model would be further improved.
The main contributions are summarized as follows:
\begin{itemize}
	\item We propose the ReDegNet to explore the real degradation from face images by explicitly learning the degradation-aware and content-independent representations which control the degraded image generation.
	\item {We transfer these real-world degradation representations to HQ natural images to generate their realistic LQ ones for supervised real restoration.}
	\item {We provide a new manner for handling intractable degraded images by learning their degradation from face regions within them, which can be used for synthesizing these types of LQ natural images for specifically fine-tuning.}
	\item {Experimental results demonstrate that our ReDegNet can well learn the degradation representations from face images and can effectively transfer to natural ones, contributing to the comparable performance on general restoration and superior performance in specific scenarios  against the SOTAs.}
\end{itemize}

\section{Related Work}
\subsection{Blind Face Restoration}
Different from the complex textures in natural images, the specific structure in face images make it feasible to well handle the real-world LQ face images~\cite{zhu2016deep,huang2017wavelet,chrysos2017deep,progressive_face_sr,Chen_2018_CVPR,chen2020progressive}. To alleviate the sensibility for the unknown degradation, reference images or component features are suggested  for guiding the blind restoration process~\cite{li2018learning,Li_2020_CVPR,Li_2020_ECCV}. Most recently, generative face prior~\cite{karras2019style,karras2020analyzing} based methods~\cite{wang2021towards,chan2020glean,Yang2021GPEN} are proposed to improve and stabilize the restoration quality, which can robustly restore the real-world LQ face images in most scenarios. Their great generalization on face images inspires us to explore the possibility of extending the restoration performance from the local region (\ie, face) to the whole image. 
%

\subsection{Degradation Estimation Based Blind Image Super-Resolution}

The real-world LQ images are mainly corrupted with unknown degradation parameters, so some works focus on estimating these degradation parameters and then apply non-blind restoration methods to recover it. 
Bell-Kligler~\etal~\cite{bell2019blind} firstly propose the image-specific KernelGAN to predict the blur kernels and feed them to ZSSR~\cite{shocher2018zero} for non-blind restoration.
Gu~\etal~\cite{gu2019blind} introduce iterative kernel correction method to estimate the blur kernel which further benefits the restoration results.
Luo~\etal~\cite{luo2020unfolding} alternate the optimization of restoring HQ images with the predicted kernel and estimating the blur kernel with the restored results, both of which can compensate each other.
Wang~\etal~\cite{wang2021unsupervised} suggest a degradation-aware super-resolution network that learn the degradation related parameters to guide the restoration process. However, real-world LQ images usually have high frequency noises or compression artifacts, and these methods are sensitive with them, which brings adverse  effect for parameter prediction.

\subsection{Data-driven  Based Blind Image Super-Resolution}
The main challenge of blind image super-resolution task can be ascribed to the lack of suitable training pairs. So a straightforward way is to collect the real-world LQ and HQ pairs. Cai~\etal~\cite{cai2019toward}  adjust the focal length of the digital cameras to capture the paired LQ/HQ images on  the same scene.  Wei~\etal~\cite{wei2020component} build a larger dataset with a large-scale diverse benchmark by zooming the digital cameras. Except for the cumbersome capturing process, the spatial and brightness misalignment easily leads to uncontrollable errors.  Moreover, although these images are realistic, they are more suitable for the specific super-resolution task that has the similar capturing scenarios. These types of collecting data occupies very few  of these complex real-world degraded images, resulting in the failure cases when handling other real degradation, \eg, noise or compression.

To alleviate the difficulties in synthesizing real-world LQ images, recent works tend to learn the restoration mapping with unpaired LQ and HQ images.
Yuan~\etal~\cite{yuan2018unsupervised} suggest a Cycle-in-Cycle network 
by firstly mapping the LQ input to noise-free space and then super-resolving it through a pre-trained super-resolution model. Similarly, Lugmayr~\etal~\cite{lugmayr2019unsupervised}  adopt the cycle consistent loss to learn a domain distribution network to generate new LQ/HQ pairs for supervised restoration. Fritsche~\etal~\cite{fritsche2019frequency} also propose the unsupervised DSGAN model to generate the degraded LQ images with the same characteristics as the original ones. To constitute more realistic LQ images, Ji~\etal~\cite{ji2020real} extract the blur kernels via KernelGAN~\cite{bell2019blind} and noise injection through~\cite{chen2018image,zhou2019kernel}, which perform on HQ images to simulate the real degradation process. Although these methods achieve great performance in most cases, they still show limited generalization ability in super-resolving real-world LQ images, because  1) the estimated degradation parameters from only a single image is highly ill-posed and they are not enough to infer how the HQ images degraded (Figure~\ref{fig:fig1} (a)), and 2) the real-world LQ images usually suffer from complex degradation, which is challenging to model due to the lack of paired data. 
In contrast, our ReDegNet adopts the pairs of real-world LQ and pseudo HQ face images to explore the real degradation process. %

Another way is to extend the degradation space. Instead of the traditional degradation process that degrades the HQ image with Gaussian blurring, followed by the bicubic downsampling operation, and the injection of Gaussian noise and JPEG compression, Zhang~\etal~\cite{zhang2021designing} propose a practical degradation model with randomly shuffled orders of these operations which tremendously cover the diverse degradation space. Similarly, Wang~\etal~\cite{wang2021realesrgan} suggest a high order degradation model with several repeated degradation process. Although these two methods show great generalization in handling real-world images, they are still incapable for those images corrupted with complex degradation like the halftone image in Figure~\ref{fig:fig1} (a).
Traditional methods remove these continuous noisy dots mainly through  filters~\cite{kite2000fast,luo1998robust,miceli1992inverse}, look-up-tables~\cite{chung2005inverse}, dictionary learning~\cite{freitas2016enhancing}, or maximum a posteriori estimation~\cite{stevenson1997inverse}.  Recent CNN-based inverse halftoning methods~\cite{xiao2017deep,xia2018deep,gao2019deep,son2020inverse} and 
even these estimation or data-driven based methods 
still fail to generate photo-realistic results on these types of real-world LQ images, which can be ascribed to the difficulties in synthesizing proper LQ images.

\section{Methodology}
Our ReDegNet aims to learn the {re}al {deg}radation from the pairs of real-world LQ and pseudo HQ face images, and transfer it to natural ones. So it mainly contains two sub-networks, \ie, DegNet for learning the {deg}radation {rep}resentation $\Omega$, and SynNet for synthesizing the LQ images  with the given 
HQ input and $\Omega$.
With the collected real-world LQ face images $I^{\textit{ReaL}}_f$ and their pseudo HQ ones $I^{\textit{PseH}}_f$, the learning process of DegNet ($\mathcal{F}_{\textit{Deg}}$) and SynNet ($\mathcal{F}_{\textit{Syn}}$) can be formulated as:
\begin{equation}
	\small
	\label{eqn:degrep}
	\Omega^{\textit{Rea}}_{f} = \mathcal{F}_{\textit{Deg}}\left( I^{\textit{ReaL}}_f, I^{\textit{PseH}}_f;\Theta_{\textit{Deg}} \right)\,,
\end{equation} 
\begin{equation}
	\small
	\label{eqn:recf}
	\hat{I}^L_f = \mathcal{F}_{\textit{Syn}}\left( I^{\textit{PseH}}_f,\Omega^{\textit{Rea}}_{f};\Theta_{\textit{Syn}} \right)\,,
\end{equation} 
where $\Theta_{Deg}$ and $\Theta_{Syn}$ are the learnable parameters for DegNet and SynNet. 

After jointly end-to-end learning through degradation disentanglement, the synthetic realistic LQ natural images can be obtained in the inference through:
\begin{equation}
	\small
	\label{eqn:recn}
	\hat{I}^L_n = \mathcal{F}_{\textit{Syn}}\left( I^H_n,\Omega^{\textit{Rea}}_{f};\Theta_{\textit{Syn}} \right),
\end{equation} 
where $I_n^H$ and $\hat{I}^L_n$ are the HQ and the synthetic LQ natural images, respectively. $\Omega^{\textit{Rea}}_{f}$ can be sampled from these real degradation representations which are extracted from the collected real-world face pairs. 
The whole framework and each sub-network are illustrated in Figure~\ref{fig:pipeline} and will be introduced in the following.

\subsection{Learning Real Degradation  from Face Image}
\begin{figure*}[!t]
	\centering
	\includegraphics[width=1.\textwidth]{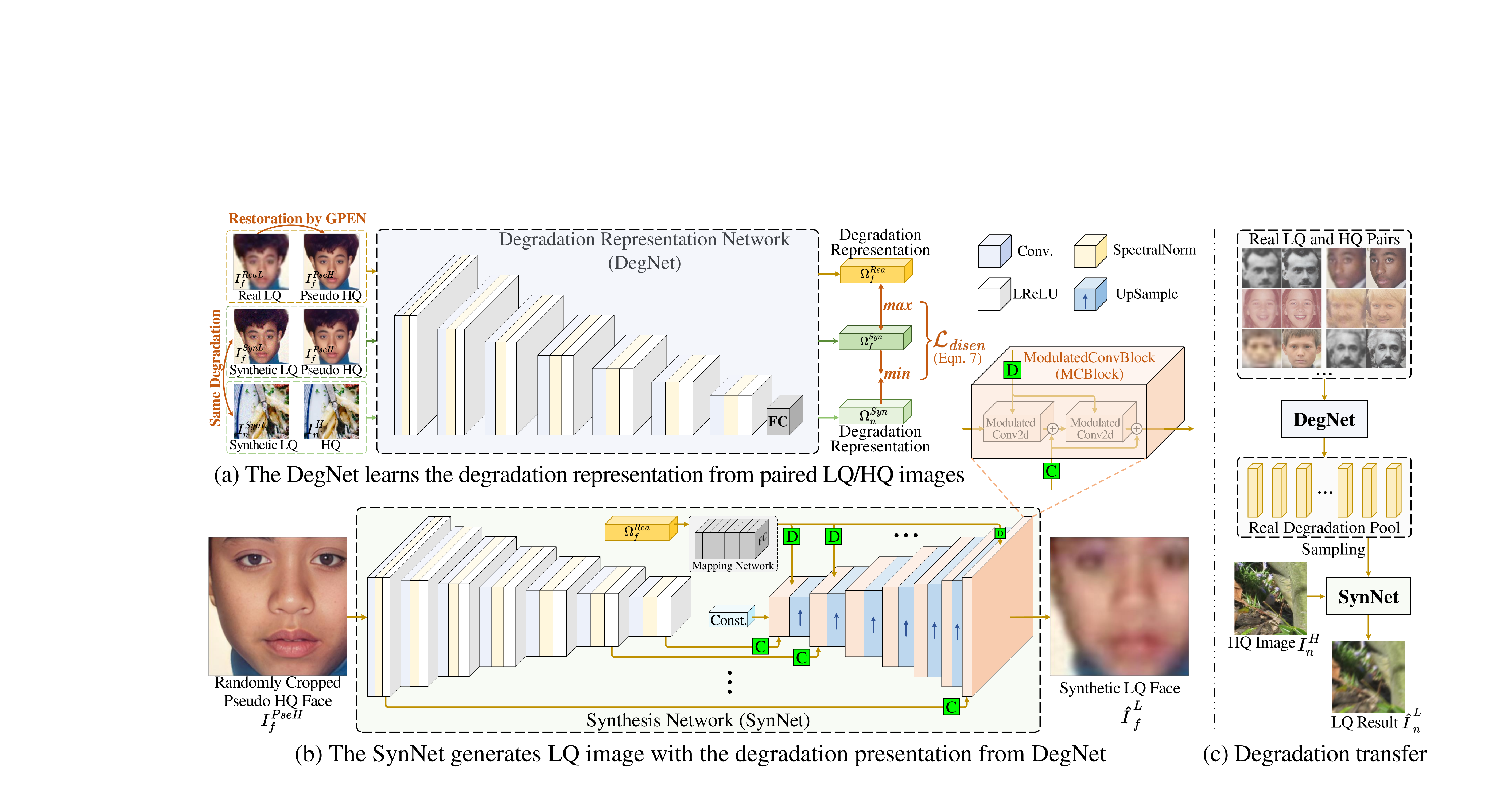}
	\caption{Overview of our ReDegNet. (a) The DegNet learns the degradation representation. (b) The SynNet synthesizes the LQ image with the degradation presentation $\Omega$ from DegNet.  {\setlength{\fboxsep}{1.5pt}\setlength{\fboxrule}{0.5pt}\fcolorbox{black}{green}{{D}}} denotes a learned affine transform from $\Omega$ that produces a degradation style. {\setlength{\fboxsep}{1.5pt}\setlength{\fboxrule}{0.5pt}\fcolorbox{black}{green}{{C}}} represents the content features that will be degraded by {\setlength{\fboxsep}{1.5pt}\setlength{\fboxrule}{0.5pt}\fcolorbox{black}{green}{{D}}} through modulated convolution. (c) The HQ natural image together with the degradation representation  sampled from the face pairs are taken into SynNet to generate their synthetic LQ one. 
	}
	\label{fig:pipeline}
\end{figure*}

Instead of predicting the degradation related representations from only a single LQ image~\cite{ji2020real,wang2021unsupervised}, 
we take the LQ and HQ pairs as input to explore the degradation process about how the HQ image is degraded to the LQ one. 
The degradation representation network (DegNet) shown in Figure~\ref{fig:pipeline}~(a)
is stacked with several convolutional layers, each of which followed by spectral normalization~\cite{miyato2018spectral} and LeakyReLU activation. A fully convolutional (FC) layer is incorporated in the last to predict the degradation representation vector $\Omega$, which has the size of $1\times512$. This sub-network is optimized through two terms, \ie, the disentanglement loss  in Eqn.~\ref{eqn:disen} and the gradient back propagated from the following SynNet.

\subsection{Synthesizing the LQ Image} 
After obtaining the degradation representation vector $\Omega$, the remaining problem is about how to utilize it to control the degradation process. Inspired by the StyleGANs~\cite{karras2019style,karras2020analyzing} that control the style of the generated image with one vector within $\mathcal{W}$ space, we adopt the similar structure to map the degradation representation $\Omega$ to $\mathcal{W}$ space through several fully convolutional (FC) layers. 
Then, instead of feeding the broadcast noise in StyleGAN, the image content of our SynNet is provided by the features of the input HQ images. 
Finally, with the degradation styles {\setlength{\fboxsep}{1.5pt}\setlength{\fboxrule}{0.5pt}\fcolorbox{black}{green}{{D}}} and image content {\setlength{\fboxsep}{1.5pt}\setlength{\fboxrule}{0.5pt}\fcolorbox{black}{green}{{C}}}, the degraded image is reconstructed with the modulated convolution operation (MCBlock) in which the degradation styles serve as the convolutional weights to control the degradation process of the given image content~\cite{karras2020analyzing}. With several cascaded MCBlocks, the final LQ result which is expected to have the similar degradation types with the given degradation representation can be synthesized. Since the degradation vector $\Omega$ should be a global representation without any spatial information, here we randomly crop the HQ image as the input of SynNet to alleviate the spatial dependency. 

To introduce different scales of textures in the training phase, we adopt the random rotation, resampling, and cropping on face images in DegNet and SynNet, simultaneously. The proposed SynNet combined with DegNet constitutes our ReDegNet that can be jointly optimized in a supervised end-to-end manner.

\subsection{Transferring Degradation to Natural Image}
After training on face images, our ReDegNet can not only extract the real degradation representation from pairs of face images, but also generate the corresponding LQ image with the expected degradation styles. So as for general restoration, we store large amounts of degradation representations that are extracted from real-world LQ  and their pseudo HQ face images, which will be sampled to imitate the real degradation process on natural HQ images (Figure~\ref{fig:pipeline} (c)). Here we also resample and rotate the LQ/HQ face pairs to augment the degradation space. Notably, our ReDegNet can be utilized in some specific restoration, in which the degradation types are not easy to synthesize with current degradation model. For the intractable old photos (\eg, Figure~\ref{fig:fig1}) or old films, we can obtain their degradation representations  with DegNet through the pairs of LQ face region within them and its pseudo HQ result. Then the HQ natural images can be utilized to generate the  corresponding LQ image by SynNet to synthesize these degradation types of natural training pairs, which can be used to fine-tune the specific restoration on the whole image. 

\subsection{Learning Objective}
Two types of loss functions are collaborated together to constrain the learning of our ReDegNet, \ie, (i) disentanglement loss that is introduced to extract the degradation-related representations, and (ii) reconstruction loss that is suggested to constrain the synthetic results close to the ground-truth.

\noindent \textbf{Disentanglement Loss.} 
The degradation representations $\Omega^{\textit{Rea}}_f$ learned from face images are expected  to perform on natural ones to control the degradation styles, so it should be degradation-aware and content-independent. To achieve this goal, we adopt contrastive learning~\cite{schroff2015facenet,wang2021unsupervised} to minimize the distances of $\Omega$s that are obtained from images with different content but have the same degradation parameters, and meanwhile maximize these negative pairs. To synthesize the degraded face and natural images with the same degradation parameters, we adopt the handcrafted degradation model from BSRGAN~\cite{zhang2021designing} to control the degradation process.
To clarify the notations, we give a unified definition $I^{\blacktriangle}_{\blacktriangledown}$ in which $\blacktriangle \in \{\textit{SynL}, \textit{ReaL}, \textit{PseH}, \textit{H}\}$ denotes the handcrafted synthetic LQ image with BSRGAN~\cite{zhang2021designing}, real-world LQ image, the restored pseudo HQ image, and real-world HQ image, respectively, $\blacktriangledown \in \{f, n\}$ represents the face and natural image, respectively. 
Denote these synthetic face and natural pairs with BSRGAN by $\{I^{\textit{PseH}}_f,I^{\textit{SynL}}_f\}$ and $\{I^H_n, I^{\textit{SynL}}_n\}$. It should be noted that $I^{\textit{SynL}}_f$ and $I^{\textit{SynL}}_n$ are obtained from $I^{\textit{PseH}}_f$ and $I^{H}_{n}$ with the same degradation sequence and parameters.
As for these three types of pairs, \ie, real-world LQ and HQ face pairs, synthetic LQ and HQ face pairs, as well as synthetic LQ and HQ natural pairs, their degradation representations can be formulated as:
\begin{equation}
	\small
	\label{eqn:q1}
	\Omega^{\textit{Rea}}_{f} = \mathcal{F}_{\textit{Deg}}\left( I^{\textit{ReaL}}_f, I^{\textit{PseH}}_f;\Theta_{\textit{Deg}} \right)\,,
\end{equation} 
\begin{equation}
	\small
	\label{eqn:q2}
	\Omega^{\textit{Syn}}_{f} = \mathcal{F}_{\textit{Deg}}\left( I^{\textit{SynL}}_f, I^{\textit{PseH}}_f;\Theta_{\textit{Deg}} \right)\,,
\end{equation} 
\begin{equation}
	\small
	\label{eqn:q3}
	\!\!\!\!\!\!\!\!\Omega^{\textit{Syn}}_{n} = \mathcal{F}_{\textit{Deg}}\left( I^{\textit{SynL}}_n, I^H_n;\Theta_{\textit{Deg}} \right)\,.
\end{equation} 
Then the disentanglement loss $\mathcal{L}_{disen}$ can be further  formulated as:
\begin{equation}
	\small
	\label{eqn:disen}
	\mathcal{L}_{\textit{disen}}=\left\|\Omega^{\textit{Syn}}_{f}-\Omega^{\textit{Syn}}_{n}\right\|_{2}^{2}+\frac{\lambda}{\|\Omega^{\textit{Syn}}_{f}-\Omega_{f}^{\textit{Rea}}\|_{2}^{2}+\epsilon}+\frac{1}{2}\left\|\Theta_{\textit{Deg}} \right\|^2_{2}\,,
\end{equation} 
where $\lambda$  is the trade-off parameter. 
By minimizing the distance between $\Omega^{\textit{Syn}}_{f}$ and $\Omega^{\textit{Syn}}_{n}$ which share the same degradation process but have the different contents (\ie, face and nature), we can constrain the extraction of degradation-aware and content-independent representations. On the contrary, by maximizing the distance between $\Omega^{\textit{Syn}}_{f}$ and $\Omega^{\textit{Rea}}_{f}$ which have the same contents (\ie, $I^{\textit{PseH}}_{f}$) but are corrupted with different degradation process, the degradation representation can be further constrained to the degradation-aware learning.

\noindent \textbf{Reconstruction Loss.} 
%
It mainly contains three terms, \ie, i) mean square error loss $\mathcal{L}_{\textit{mse}}$, ii) realistic loss $\mathcal{L}_{real}$, and iii) degradation-consistent loss $\mathcal{L}_{cons}$.

i) The MSE loss $\mathcal{L}_{\textit{mse}}$ contains two terms and  is formulated as:
\begin{equation}
	\small
	\mathcal{L}_{\textit{mse}}\!=\!\ell_{\textit{mse}}(\hat{I}^L_f, I^{\textit{ReaL}}_f)
	\!=\!\frac{1}{\mathcal{{CHW}}}\left\|\hat{I}^{L}_f\!-\!I^{\textit{ReaL}}_f\right\|^2\!\!+\!\sum_{i=1}^{4} \!\frac{0.1}{\mathcal{C}_{i} \mathcal{H}_{i} \mathcal{W}_{i}}\left\|\Phi_{i}(\hat{I}^{L}_f)\!-\!\Phi_{i}(I^{\textit{ReaL}}_f)\right\|^{2}\!\!
	\label{eqn:mse}
\end{equation}
where $\hat{I}^L_f$ is the generated LQ face image in Eqn.~\ref{eqn:recf} and $I^{\textit{ReaL}}_f$ is the collected real-world LQ image. $\mathcal{C}_*$, $\mathcal{H}_*$, $\mathcal{W}_*$ are the dimensions and $\Phi_{i}$ is the $i$-th
convolution layer of the pre-trained VGG-19 model~\cite{simonyan2014very}. This objective constrains the synthetic LQ images close to the real-world LQ images in both pixel and feature space~\cite{johnson2016perceptual}.

ii) The realistic loss $\mathcal{L}_{real}$ mainly considers two types of constraints, \ie, style loss~\cite{gatys2016image} and adversarial loss~\cite{goodfellow2014generative}. The first one is computed with the Gram matrix on the feature spaces of VGG-19 model 
and can be formulated as:
\begin{equation}
	\small
	\mathcal{L}_{\textit{style}}=\sum_{i=1}^{4}\frac{1}{\mathcal{C}_{i} \mathcal{H}_{i}\mathcal{W}_{i}}\left\|\Phi_{i}(\hat{I}^{L}_f)^{\!T}\Phi_{i}(\hat{I}^{L}_f)-\Phi_{i}(I^{\textit{ReaL}}_f)^{\!T}\Phi_{i}(I^{\textit{ReaL}}_f)\right\|^{2}\,,
\end{equation}
in which the variants have the same definitions as these in Eqn.~\ref{eqn:mse}. The second one is the widely used adversarial loss which is effective in constraining the results within the natural manifold. In this paper, we adopt the discriminator from SNGAN ~\cite{miyato2018spectral} by incorporating the spectral normalization behind each convolutional layer.
It is worth noting that the result $\hat{I}^L_f$ is expected to be a LQ image and visually blur in most cases, which is difficult for discriminator to distinguish whether it is a real LQ or fake LQ image due to the wider space of LQ types. So instead of only taking the synthetic result into the discriminator, we take the HQ image and their degradation representation as additional conditions~\cite{mirza2014conditional}.
The hinge version of adversarial loss~\cite{miyato2018spectral,zhang2019self} is given by:
\begin{gather}
	\small
	\!\!\mathcal{L}_D\!\!=\!\!-\mathbb{E}[\min(0,\!\!-1\!\!+\!\!D(I^{\textit{ReaL}}_f\!\!, I^{\textit{PseH}}_f\!\!,\Omega^{\textit{Rea}}_f))]
	\!-\!\mathbb{E}[\min(0,\!\!-1\!\!-\!\!D(\hat{I}^L_f\!, I^{\textit{PseH}}_f\!\!, \Omega^{\textit{Rea}}_{f}))]\!\\
	\mathcal{L}_G=-\mathbb{E}[D(\mathcal{F}_{\textit{Syn}}(I^{\textit{PseH}}_{f},\mathcal{F}_{\textit{Deg}}(I^{\textit{ReaL}}_f, I^{\textit{PseH}}_f; \Theta_{\textit{Deg}});\Theta_{\textit{Syn}}), I^{\textit{PseH}}_f,\Omega^{\textit{Rea}}_{f})]\,.
\end{gather}
Combining the two terms together, the final realistic loss is formulated as:
\begin{equation}
	\small
	\setlength{\abovedisplayskip}{5pt}
	\setlength{\belowdisplayskip}{5pt}
	\mathcal{L}_{\textit{real}} = 0.1\cdot\mathcal{L}_{\textit{style}}+\mathcal{L}_{G}\,.
\end{equation}

iii) The third one is the degradation-consistent loss.
As analyzed before, the degradation representation $\Omega^{Syn}_{f}$ and $\Omega^{Syn}_{n}$ in Eqns.~\ref{eqn:q2} and \ref{eqn:q3} are obtained from the face and natural pairs that are corrupted by the same degradation process. 
Therefore, switching $\Omega^{Syn}_f$  and $\Omega^{Syn}_n$ should have the same LQ results.
Thus the degradation-consistent loss is suggested as:
\begin{equation}
	\small
	\mathcal{L}_{\textit{cons}}=\ell_{\textit{mse}}(\mathcal{F}_{\textit{Syn}}(I^H_n,\Omega^{\textit{Syn}}_{f}; \Theta_{\textit{Syn}}), I^{\textit{SynL}}_n) +
	\ell_{\textit{mse}}(\mathcal{F}_{\textit{Syn}}(I^{\textit{PseH}}_f,\Omega^{\textit{Syn}}_{n}; \Theta_{\textit{Syn}}), I^{\textit{SynL}}_f)\,,
\end{equation}
where $\ell_{\textit{mse}}$ is the MSE loss defined in Eqn.~\ref{eqn:mse}. With the constraints on the degradation representation $\Omega^{\textit{Syn}}_{f}$ ($\Omega^{\textit{Syn}}_{n}$) that is extracted from face (natural) images and performed on natural (face) ones, we can further optimize the disentanglement learning, and benefit the training process of  the SynNet. 

To sum up, the final learning objective is formulated as:
\begin{equation}
	\small
	\mathcal{L}=\lambda_{\textit{disen}}\mathcal{L}_{\textit{disen}}+\lambda_{\textit{mse}}\mathcal{L}_{\textit{mse}}+\lambda_{\textit{real}}\mathcal{L}_{\textit{real}}+ \lambda_{\textit{cons}}\mathcal{L}_{\textit{cons}}\,,
\end{equation}
where $\lambda_{\textit{disen}}$, $\lambda_{\textit{mse}}$, $\lambda_{\textit{real}}$ and $\lambda_{\textit{cons}}$ are set to 5, 1, 0.1, and 2, respectively.

\section{Experiments}
Since our ReDegNet is proposed to design a degradation model for synthesizing LQ images, in this work, we mainly compare  with three related works, \ie, RealSR~\cite{ji2020real}, BSRGAN~\cite{zhang2021designing} and Real-ESRGAN~\cite{wang2021realesrgan}, in which RealSR synthesizes the LQ image with the estimated kernel and noise from the single real-world photograph, BSRGAN and Real-ESRGAN focus on handcrafted design of diverse degradation. These three methods and Ours adopt the same network (\ie, ESRGAN~\cite{wang2018esrgan}) 
, so we can fairly compare with their released models. To evaluate the effectiveness of blind image super-resolution methods on handling the real-world LQ images, here we analyze the performance on two types of real-world images, \ie, real-world pairs collected by digital camera, and real-world single LQ images. As for the quantitative evaluation, we use PSNR, SSIM, and LPIPS~\cite{zhang2018perceptual} to measure the distance between the result and ground-truth.  Since real-world single LQ images do not have the ground-truth, we follow the competing methods~\cite{ji2020real,zhang2021designing,wang2021realesrgan} and adopt NIQE~\cite{mittal2012making} to evaluate the non-reference image quality. %

\subsection{Dataset and Implementation Details}
We collect real-world LQ face images from Internet, and 
then adopt GPEN~\cite{Yang2021GPEN} to  obtain their pseudo HQ counterparts. These images cover diverse degradation types, from slightness to severeness, oldness to present, 
\etc. Among them, 10,000 images are used for training, 1,000 images for validating, and the remaining 5,000 images for testing. 
Except these collected images,
we also introduce the synthetic LQ face images from FFHQ~\cite{karras2019style} with common degradation, \eg, blur, noise,  JPEG compression, and downsampling operation, \etc, to improve the generalization ability. During the inference, we conduct the degradation representation pool $\{\Omega^{ReaL}_{f}\}^{N}$ from these face pairs, which will be sampled to constitute the natural pairs for training our general restoration network (\ie, F2N-ESRGAN).

As for the natural image, we follow BSRGAN~\cite{zhang2021designing}, and adopt DIV2K~\cite{Agustsson_2017_CVPR_Workshops}, Flick2K~\cite{lim2017enhanced,timofte2017ntire} and FFHQ~\cite{karras2019style} for training our ReDegNet and F2N-ESRGAN.
%
Adam optimizer~\cite{kingma2014adam} with $\beta_1=0.5$ and $\beta_2=0.999$ is adopted to train ReDegNet and F2N-ESRGAN.  The initial learning rate is set to $2\times10^{-4}$ and will decrease by 0.5 when the MSE loss $\mathcal{L}_{mse}$ on the validation set tends to be stable. All the experiments are implemented on a PC server with 4 Tesla V100 GPUs. 
%

\subsection{Quantitative Comparison}
Table~\ref{tab:com} lists the quantitative results. One can see that (i) as for these real-world pairs (RealSR Canon and Nikon~\cite{cai2019toward}, and DRealSR~\cite{wei2020component}), although the PSNR and SSIM of Ours is comparable  against others, the LPIPS of Ours obtains the best performance, which indicates that our results are more consistent with human perception~\cite{zhang2018perceptual}. 
The best LPIPS of Ours in turn validates  the effectiveness of our ReDegNet in synthesizing the realistic training pairs. 
(ii) As for the non-reference image quality metric, we collect two groups of real-world images, \ie, RealSRSet proposed in BSRGAN~\cite{zhang2021designing}, and RealLQSet that contains LQ images collected from Internet and LQ frames extracted from 480P videos. We can see that results of Ours are better than others in most cases, but inferior to RealSR~\cite{ji2020real} in RealSRSet~\cite{zhang2021designing}. We analyze that the RealLQSet (1,000 images) covers more types of common real-world LQ images than RealSRSet (only 20 images), which indicates RealLQSet is more suitable in evaluating the performance of super-resolving the real-world LQ images. The better NIQE of Ours may be attributed to the usage of degradation that are learned from real-world LQ face images.
%

\begin{table*}[t]
	\centering
	\scriptsize
	\renewcommand\arraystretch{1.1}
	\caption{{Quantitative comparison on two types of real-world LQ images.}}
	\setlength{\tabcolsep}{0.35mm}
	{
		\begin{tabular}{|c| c c c| c c c| c c c|cc|}
			\hline
			\rowcolor{lightgray}\rowcolor{lightgray}
			\rowcolor{lightgray}&\multicolumn{9}{c|}{Real-world Pairs}&
			\multicolumn{2}{c|}{Real-world LQ}\\
			\hhline{>{\arrayrulecolor{lightgray}}-|>{\arrayrulecolor{black}}-----------}
			\rowcolor{lightgray}& \multicolumn{3}{c|}{RealSR-Canon} & \multicolumn{3}{c|}{RealSR-Nikon}& 
			\multicolumn{3}{c|}{DRealSR} &
			\multicolumn{1}{c}{RealSRSet} & \multicolumn{1}{c|}{RealLQSet}\\
			\rowcolor{lightgray}\multirow{-3}{*}{\makecell[c]{\textbf{Methods}}}&
			\tiny PSNR$\uparrow$ &\tiny SSIM$\uparrow$ &\tiny  LPIPS$\downarrow$ &\tiny  PSNR$\uparrow$ &\tiny SSIM$\uparrow$ &\tiny  LPIPS$\downarrow$&\tiny
			PSNR$\uparrow$ &\tiny SSIM$\uparrow$ &\tiny  LPIPS$\downarrow$ &\tiny
			NIQE$\downarrow$ &\tiny NIQE$\downarrow$\\
			\hline \hline
			RealSR& \underline{25.58}  & .723 & .458   & \textbf{25.49}  & .693 & .459 & \textbf{27.69}  & \textbf{.759} & .438   & \textbf{4.82} & 5.62  \\
			BSRGAN&  \textbf{25.61}  & \textbf{.768} & \underline{.363}   & 24.51 & \underline{.711} & .391 & 26.64  & .744 & .380   & 5.60  & 5.36  \\
			Real-ESRGAN&   24.95 & \textbf{.768} & .366   & 24.50 & \textbf{.716} & \underline{.388} &  26.57 & .753 & \underline{.374}   & 5.75 & \underline{5.24}  \\
			Ours& 25.57  & \underline{.765} & \textbf{.362}   & \underline{25.43}  & \textbf{.716} & \textbf{.385} & \underline{26.91}  & \underline{.758}  & \textbf{.373}   & \underline{4.85} & \textbf{4.93}  \\
			\hline
			Ours (\textit{-}$D$)& 24.63  & {.749} & {.463}   & {24.35}  & {.684} & {.460} & {26.32}  & {.740}  & {.425}   & {6.45} & {6.27}  \\
			Ours (\textit{U})& 25.05  & {.752} & {.428}   & {24.72}  & {.708} & {.421} & {26.35}  & {.741}  & {.404}   & {5.81} &{5.93}  \\
			\hline
	\end{tabular}}
	\label{tab:com}
\end{table*}

\begin{figure*}[!t]
	\centering
	\includegraphics[width=1\textwidth]{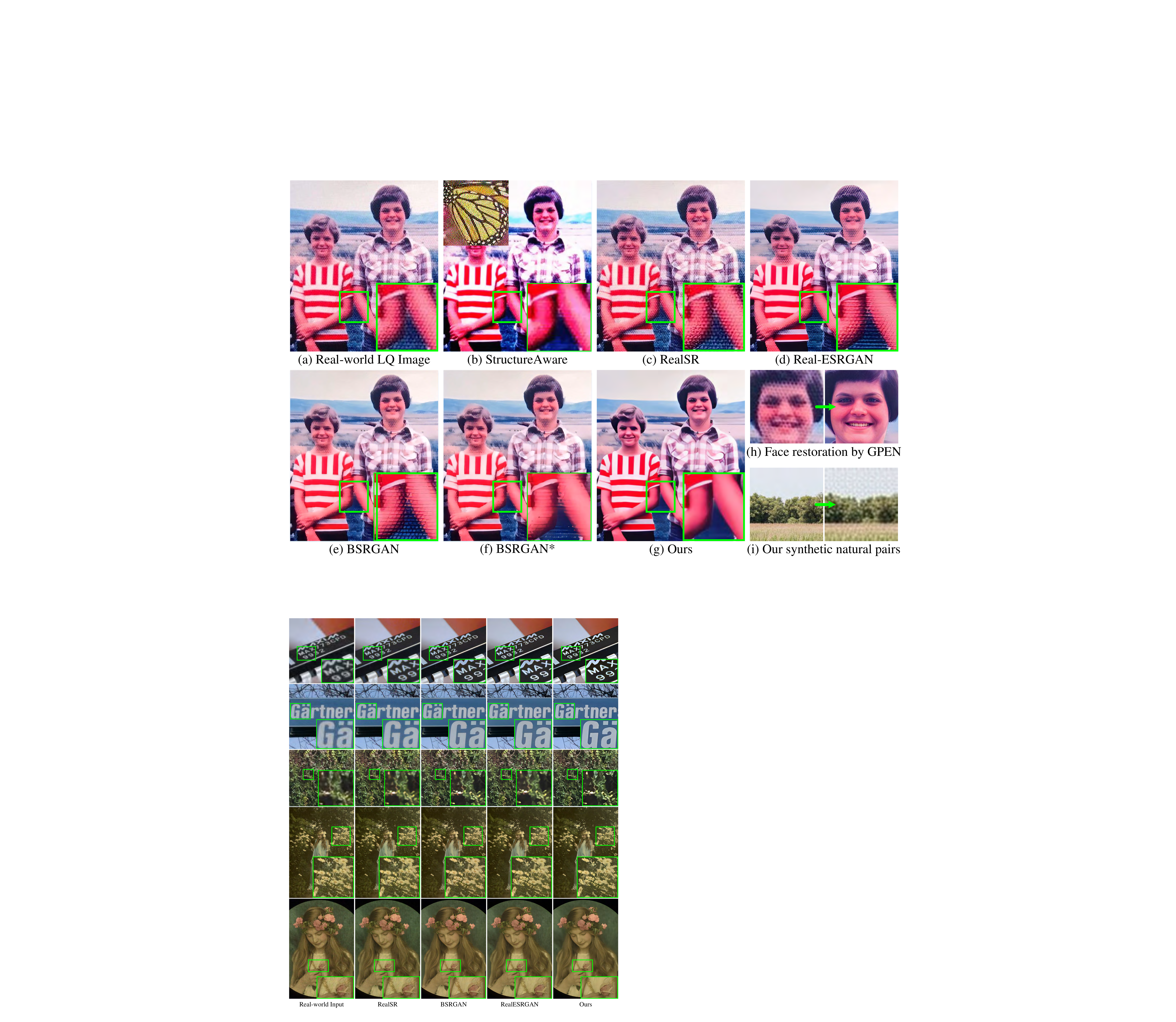}
	\caption{Visual comparison of these competing methods on real-world LQ images. 
	}
	\label{fig:viscom}
\end{figure*}

\begin{figure*}[!t]
	\centering
	\includegraphics[width=1\textwidth]{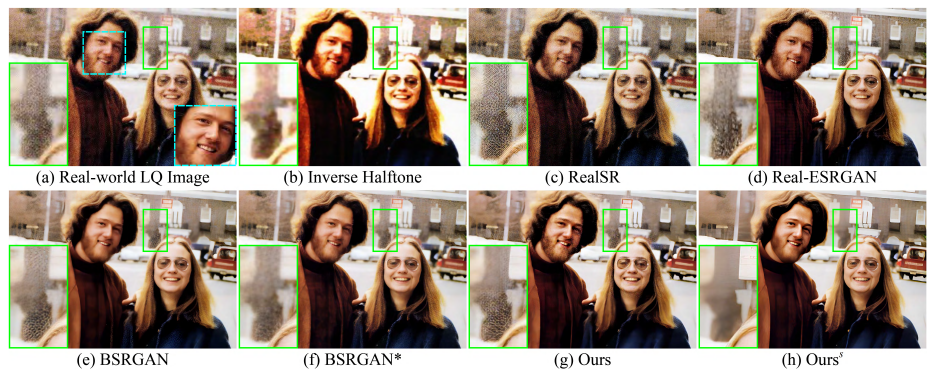}
	\caption{Restoration results on real-world LQ image. \textcolor[rgb]{0,1,1}{Close-up} on the right bottom of (a) is the face restoration result by GPEN~\cite{Yang2021GPEN}. BSRGAN* denotes the official BSRGAN model fine-tuned with the incorporation of halftone degradation~\cite{gao2019deep}. Ours$^s$ represents our general model (Ours) that is specifically fine-tuned with the synthetic natural pairs with the degradation representation from the face region. 
	Best view it by zooming in.
	}
	\label{fig:spe}
\end{figure*}
\subsection{Visual Comparison on Real-world  LQ Images}
Except the quantitative metrics, visual comparison appears to be critically important in evaluating the restoration performance, especially for these real-world LQ images. In this paper, we select these real-world LQ images from three  types of datasets, \ie, RealSRSet from BSRGAN~\cite{zhang2021designing}, RealSR dataset~\cite{cai2019toward}, and the collected real-world LQ images from our RealLQSet. Visual results of the competing methods are shown in Figure~\ref{fig:viscom}. One can see that results of Ours are much clearer than others, not only in the smooth regions (1\textit{st} row), but also in these with rich and complex textures (2$\sim$4\textit{th} rows). 
Due to the limited ability in predicting the kernel and noise from the real-world LQ images, RealSR~\cite{ji2020real} fails to generate plausible and photo-realistic textures when handling the input with complex degradation. 
Although BSRGAN~\cite{zhang2021designing} and Real-ESRGAN~\cite{wang2021realesrgan} show great generalization due to the wider handcrafted degradation spaces, our F2N-ESRGAN performs comparable against them with these degradation representations that are extracted from the real-world LQ face images, which indicates the effectiveness of our method in synthesizing the  photo-realistic LQ images, and in turn contributes to the better restoration performance.

\subsection{Fine-tuning for Specific Restoration}
Except the general super-resolution task mentioned above, our method can also fine-tune the restoration model on specific scenarios which have face images in them. Figures~\ref{fig:fig1} and \ref{fig:spe} show the specific cases. We can observe that (1) although they are similar to the halftone degradation, the restoration result by the inverse halftone method~\cite{son2020inverse} can not well handle it (see (b)) due to the complex degradation that these real-world LQ images usually suffer from.
(2) The general restoration methods, \ie, RealSR~\cite{ji2020real}, BSRGAN~\cite{zhang2021designing}, and Real-ESRGAN~\cite{wang2021realesrgan} also fail to generate plausible results on these unsynthesizable degradation  (see (c$\sim$e)), while Ours perform favorable but still contain obvious artifacts (see (g)). (3) By fine-tuning BSRGAN with the synthetic halftone degradation~\cite{gao2019deep}, BSRGAN* has slight improvement in reducing the artifacts, but still can not generate photo-realistic structures (see (f)). (4) By restoring the face region with GPEN~\cite{Yang2021GPEN} and synthesizing the similar degradation types on natural images (see our suppl.), results of Ours$^s$ are much better than others, which indicates 
the effectiveness of our method in learning the degradation from face images and transferring to natural ones.
Compared with BSRGAN and Real-ESRGAN, our method can not only handle the general restoration with limited real degradation, but also fine-tune the model for some specific scenarios that have face images, which are common in the consumer photography and old photos or films. 

\begin{figure}[!t]
	\centering
	\includegraphics[width=1\textwidth]{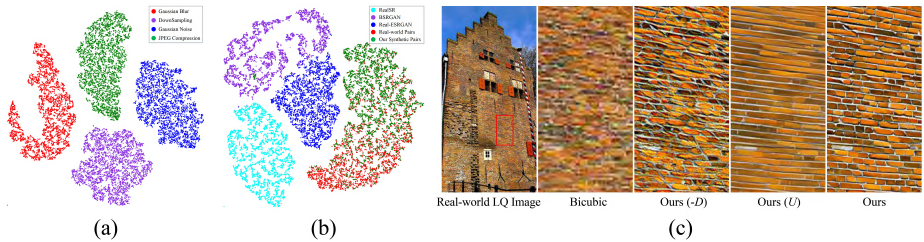}
	\caption{(a) The t-SNE results of four groups degradation with only blur, downsampling, noise and JPEG compression. (b) The t-SNE results of the synthetic degradation by the competing methods. (c) Restoration comparison of different variants.}
	\label{fig:aba}
\end{figure}

\subsection{Ablation Study}

Firstly, to illustrate the degradation extraction ability of our DegNet, we introduce t-SNE~\cite{van2008visualizing} to visualize the degradation representation $\Omega$ for different degradation types. To this end, we generate four groups of LQ face pairs by separately degrading 5,000 HQ test images with Gaussian blurring, downsampling, Gaussian noise and JPEG compression. Then DegNet is utilized to extract their degradation representations. The visualization of each group mapping to 2D space by t-SNE is shown in Figure~\ref{fig:aba}~(a). We can observe that these four groups of degradation representations are embedded into four clusters completely, which indicates that our DegNet can well capture and distinguish the different degradation types. 

Secondly, we explore the degradation space of these competing methods. With the 5,000 real-world test LQ and HQ face pairs, we synthesize the LQ images by utilizing the degradation models of RealSR~\cite{ji2020real}, BSRGAN~\cite{zhang2021designing}, Real-ESRGAN~\cite{wang2021realesrgan} and our ReDegNet on the pseudo HQ images. Among them, the kernel and noise of RealSR are extracted from the real-world images. BSRGAN and Real-ESRGAN are used with their default settings from their released models. As for ours, we randomly sample from the degradation representation pool $\{\Omega^{ReaL}_{f}\}^{N}$ via SynNet to generate the LQ images. Note that $\{\Omega^{ReaL}_{f}\}^{N}$ have no overlap with the 5,000 test pairs. The visualization of the degradation representations of these five groups, \ie, RealSR, BSRGAN, Real-ESRGAN, Ours and the real-world LQ/HQ pairs, is shown in Figure~\ref{fig:aba}~(b). One can see that our synthetic LQ images are more consistent with the real-world LQ ones than the competing methods, indicating the effectiveness of our method in extracting the degradation from real pairs of face images. Albeit BSRGAN and Real-ESRGAN have more diversities due to the random/high orders and handcrafted degradation, only few LQ images are similar to the real-world LQ ones within 5,000 pairs.

Finally, to evaluate the necessities of the disentanglement loss and our pairs of LQ and HQ face images, we design two variants, \ie, 
Ours $(U)$ by using unpaired data which feeds only the LQ images into DegNet and random HQ images into SynNet, respectively, and adopts the discriminator to distinguish whether the result has the similar degradation with the LQ input or not, 
and Ours (\textit{-}$D$) by removing the disentanglement loss. The comparisons on real-world LQ images are shown in Table~\ref{tab:com} and Figure~\ref{fig:aba} (c). We can see that compared with Ours $(U)$, results of Ours are clearer and more photo-realistic, indicating the effectiveness of our supervised manner in predicting the real degradation from the pairs of face images. Besides, by removing the disentanglement loss, results of Ours (\textit{-}$D$) easily have distorted structures and obvious artifacts, which may be caused by the inaccurate degradation representation that may contain the face related content.

\subsection{Limitations}
This work is intuitively motivated by the observation that the face region usually shares the similar degradation with the non-face region. However, the background is sometimes out of the depth of field, which easily has the inconsistent degradation with face region, thereby bring limited benefits for the specific restoration. 
Besides, our general restoration model performs not obviously superior to the competing methods, especially on these camera captured test sets in Table~\ref{tab:com}. It is better to collect face images under the similar scenarios to augment the degradation space.

\section{Conclusion}
In this work, we made the first attempt to model the real degradation from the real-world LQ face images and their pseudo HQ counterparts, and transfer these real degradation processes to HQ natural images by disentangling the degradation-aware and content-independent representations. With the synthetic natural image pairs generated by our ReDegNet, the trained blind image super-resolution models (\ie, F2N-ESRGAN) demonstrated competitive performance against SOTA methods, especially on real-world LQ images. Our method provided a new solution to synthesize more realistic LQ natural images with the degradation representation that are extracted from the facial regions within them, which are beneficial for restoring the details of non-facial regions. Experiments showed that our ReDegNet can well learn the real degradation from face images, and can effectively generate the photo-realistic LQ natural ones, thereby leading to promising performance in general and specific restoration.

\vspace{10pt}
\noindent\textbf{Acknowledgment}
This work is partially supported by the National Natural Science Foundation of China under grant No. U19A2073, the Major Key Project of PCL under grant No. PCL2021A12,  and Hong Kong RGC RIF grant (R5001-18).

\bibliographystyle{splncs04}
\bibliography{egbib}
\end{document}